# CLUSTERING ASSISTED FUNDAMENTAL MATRIX ESTIMATION

## Hao Wu and Yi Wan

School of Information Science and Engineering, Lanzhou University, China
wuhao11905@163.com, wanyijs@163.com

## ABSTRACT

*In computer vision, the estimation of the fundamental matrix is a basic problem that has been extensively studied. The accuracy of the estimation imposes a significant influence on subsequent tasks such as the camera trajectory determination and 3D reconstruction. In this paper we propose a new method for fundamental matrix estimation that makes use of clustering a group of 4D vectors. The key insight is the observation that among the 4D vectors constructed from matching pairs of points obtained from the SIFT algorithm, well-defined cluster points tend to be reliable inliers suitable for fundamental matrix estimation. Based on this, we utilizes a recently proposed efficient clustering method through density peaks seeking and propose a new clustering assisted method. Experimental results show that the proposed algorithm is faster and more accurate than currently commonly used methods.*

## 1. INTRODUCTION

Fundamental matrix (F-matrix) is an important tool often used in many computer vision tasks. It reflects the corresponding relationship between two pictures shot at the same scene but taken from different viewpoints. It is widely used in tasks such as camera tracking, image rectification, and 3D reconstruction (e.g., [1], [2]). The fundamental matrix constrains the coordinates between two 2D images from two different viewpoints. It contains information about the cameras' focal lengths, the positions of optical centers, and the rotation and translation between the two cameras [1].

In order to calculate the F-matrix, using the fact that it contains 7 free variables, Hartley used 7 pairs of corresponding points, calculated 3 possible F-matrices, and called it 7-Point algorithm [2]. Tsai and Huang proposed a linear algorithm [3], called the 8-Point algorithm for computing the F-matrix. It uses 8 pairs of matching points and estimates the F-matrix by solving a set of linear equations and making the F-matrix subject to a rank-2 constraint. There are also some other nonlinear methods. For example, Mohr and others assumed that the errors of different corresponding points are mutually independent and have the same standard deviation. They calculated the spatial position of the 3D points and the F-matrix by minimizing the error between the observed values and the fitted values of projected 2D coordinates from the 3D points [4].

A problem with current methods is that there are usually noise (error of 1 or 2 pixels) and point pairs that are wrongly matched. But all the algorithms mentioned above need to be calculated by the least square method, which requires that the noise mean is 0, and that all matched data points are correct when estimating the F-matrix. Thus the above mentioned algorithms are susceptible to noise and incorrectly matched point pairs.

Some robust algorithms have been proposed to solve these problems. Zhang put forward an algorithm called M-Estimators that can distinguish inliers from outliers. This algorithm could adapt to noise but not the wrongly-matched points [5]. Rousseeuw and Leroy proposed a Least Median of Squares (LMeDS) algorithm. It randomly selects 7 pairs of corresponding points to calculate the fundamental matrix and the error function, then uses an adaptive threshold related to error function to find the outliers. The result is shown to be quite encouraging [6].

## 2. PROPOSED ALGORITHM

In this section we propose a new F-matrix computation method. We briefly review some preliminary topics in the first three subsections for completeness, then uses another three subsections to present the proposed algorithm.

### 2.1. Epipolar Geometry and Fundamental Matrix

Epipolar geometry describes the relationship between two images taken by two cameras $C$ and $C'$ pointing at the same scene. It is independent of the structure of the scene and is only related to the intrinsic and extrinsic parameters of the cameras.

As illustrated in Fig. 1, $I$ and $I'$ are the image planes of camera $C$ and $C'$ respectively. Their coordinate systems are set up as is commonly done. Specifically, origin point is located at the top-left of the image while the x-axis points to the right and the y-axis points to the bottom. The points $m = (x, y, 1)^T$ and $m' = (x', y', 1)^T$ are the image points in plane $I$ and $I'$ of the same 3D point $M$. Because $C$, $M$ and $m$ are on a line while $C'$, $M$, $m'$ are on another line, the six points are on a plane $\pi$, which is commonly called the epipolar plane. The image point of $C$ on the plane $I'$ is called the epipole $e'$ while $e$ is the image point of $C'$ on the plane $I$. $e$ and $e'$ are also on the plane $\pi$. The line $l : ax + by + c = 0$ which crosses $m$, $e$ and the line $l' : a'x + b'y + c' = 0$ are called the epipolar lines.

The fundamental matrix $F$ is a 3*3 matrix with rank 2. It satisfies the following constraint [1]

$$m'^T Fm = 0, \qquad\qquad\qquad\qquad (1)$$

which means the point $m'$ lies on the epipolar line $l' = Fm$ while the point $m$ lies on the epipolar line $l = F^T m'$ as well.

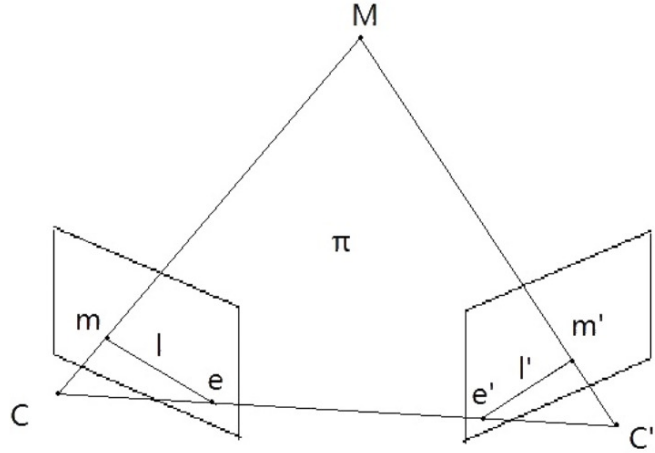

Figure 1.  Illustration of the epipolar geometry.

## 2.2. Review of the 8-Point Algorithm

Suppose the F-matrix takes the form of

$$F = \begin{pmatrix} f_{00} & f_{01} & f_{02} \\ f_{10} & f_{11} & f_{12} \\ f_{20} & f_{21} & f_{22} \end{pmatrix}. \tag{2}$$

Also suppose that at least eight pairs of points $m_n = (x_n, y_n, 1)^T$ and $m'_n = (x'_n, y'_n, 1), n = 1, 2, ..., N$ with $N \geq 8$ are given.

Then from (1), we obtain the following set of linear equations

$$Af = (a_1 a_2 ... a_N)^T f = 0, \tag{3}$$

In which

$$a_n = (x_n x'_n, y_n x'_n, x'_n, x_n y'_n, y_n y'_n, y'_n, x_n, y_n, 1)^T, \tag{4}$$

$$f = (f_{00}, f_{01}, f_{02}, f_{10}, f_{11}, f_{12}, f_{20}, f_{21}, f_{22})^T. \tag{5}$$

Without regard to the rank 2 constraint tentatively. When the F-matrix is multiplied by a scalar, it still satisfies (1). Now the F-matrix has 8 independent variables. We can utilize the constraint [3]

$$(f_{00})^2 + (f_{01})^2 + ... + (f_{22})^2 = 1. \tag{6}$$

After solving the Eq. 2, the F-matrix still does not yield to the rank 2 constraint, so we can use the SVD decomposition [3]

$$F = UDV^T = U \begin{pmatrix} \sigma_0 & 0 & 0 \\ 0 & \sigma_1 & 0 \\ 0 & 0 & \sigma_2 \end{pmatrix} V^T, \sigma_0..\sigma_1..\sigma_2..0. \tag{7}$$

Since $F$ is of rank 2, we have $\sigma_2 = 0$. Then the final estimate $F'$ of the 8-point algorithm is given as

$$F' = U \begin{pmatrix} \sigma_0 & 0 & 0 \\ 0 & \sigma_1 & 0 \\ 0 & 0 & 0 \end{pmatrix} V^T. \tag{8}$$

## 2.3. Review of RANSAC

Given any statistical model, Random Sample Consensus (RANSAC) distinguishes the outliers from a data set by using a threshold of error and then finds the best parameters which describes the data set well.

Before running RANSAC to estimate the F-matrix, we need a given *th* threshold (which is the distance between a point and its epipolar line) and the confidence $p$ of the model. Then the number of iterations $k$ and the number of inliers $d$ must satisfy $1 - p = (1 - (d/N)^8)^k$ [9]. The process is shown in Fig. 2.

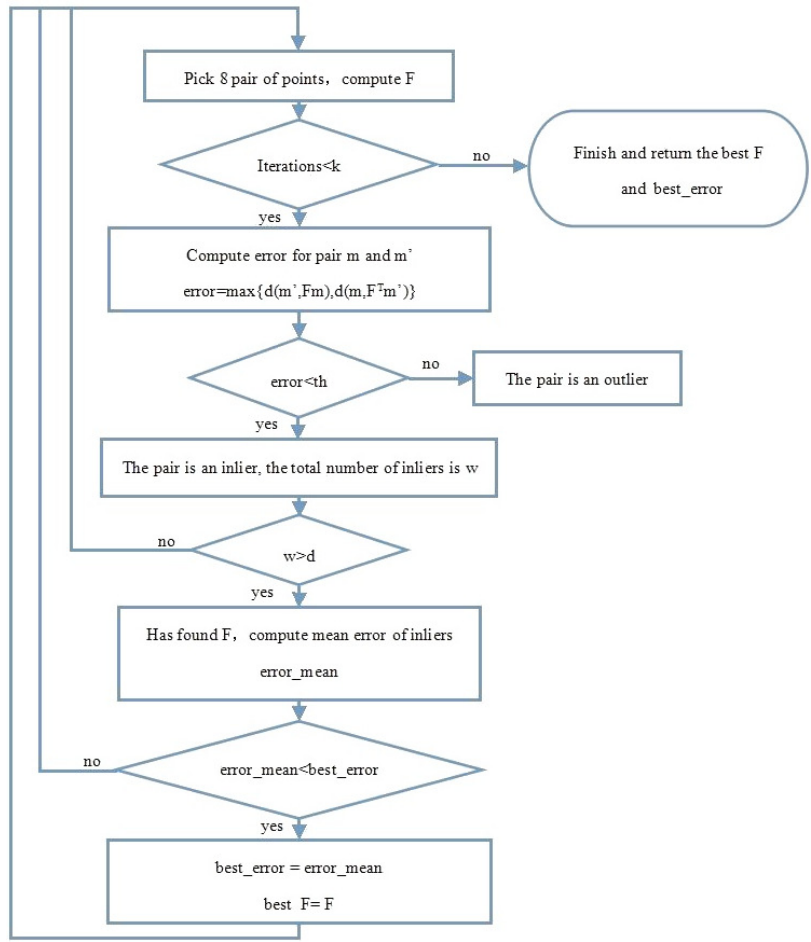

Figure 2.  Flowchart of the general RANSAC algorithm

### 2.4. The 4D Space of Matching Points

For any pair of feature points $m = (x, y, 1)^T$ and $m' = (x', y', 1)^T$, the coordinates of the points describe the positions of the points in the image planes $I$ and $I'$. The difference between the two points describes the transformation between the two viewpoints. Generally, in an image, the feature rich areas tends to have more concentrated feature points to form the matching pairs for F-matrix estimation. In order to study the distribution and transformation of the matching points, we construct a 4D space in which the vectors are of the form $q_n = (x_n, y_n, x_{n'}, y_{n'})^T$. The 4D space contains the information about both the 2D feature point distribution and the transformation between the matching points. It can help find useful characteristics of the matching points to estimate the F-matrix.

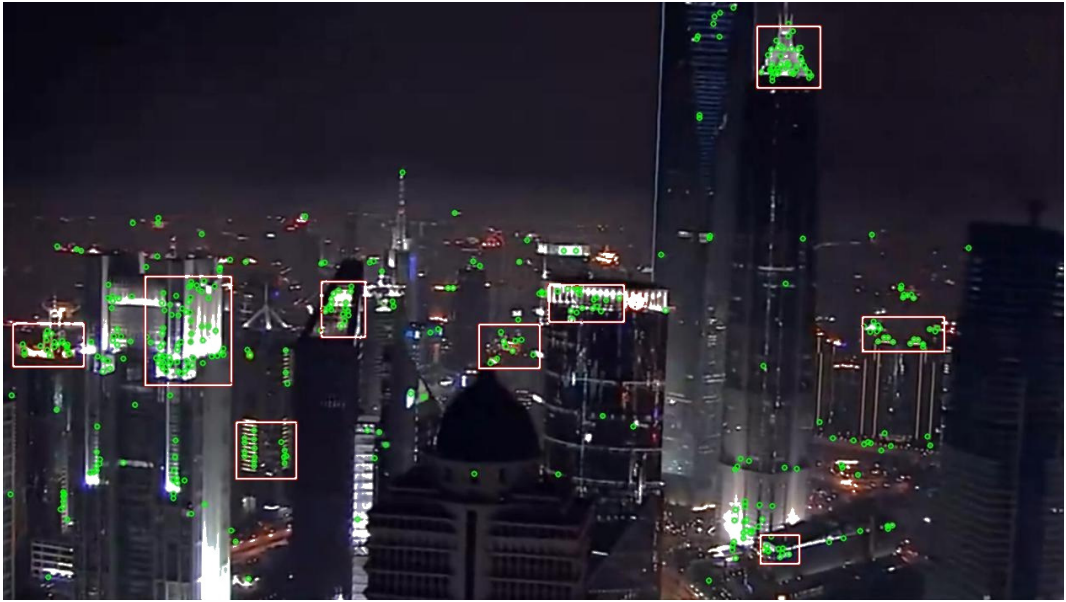

Figure 3. Feature point distribution illustration. The feature points are found by SIFT. Areas with more concentrated features points are marked out in the boxes.

Fig. 3 shows an example of the distribution of the feature points in a general natural image. After using SIFT to find the feature points in a real image pair, we get $N = 411$ pairs of matching points. We can see from the Fig. 3 that most of the feature points locate in some small areas shown in the boxes. This observation provides us with the hint that by doing a cluster analysis on this 4D vectors, we may be able to identify those feature point pairs that are more reliable than those isolated ones. In the next subsection we describe a practically efficient clustering method.

### 2.5. Cluster Analysis

Clustering analysis that classifies the data is a common method used to analyze multidimensional data. Data points that are dense in a multidimensional area and far from other dense area tend to belong to the same cluster. The clustering result describes the distribution of the data in multidimensional space. When clustering, we commonly treat the point that is closest to the center of an area as the center of the cluster. The center point of a cluster decides the basic property of the cluster. These ideas are recently used by Rodriguez and Laio in their paper "Clustering by fast search and find of density peaks". The key steps are described as follows.

(1) Computation of the local density $\rho_i$.

$$\rho_i = \sum_j \chi(d_{ij} - d_c),$$ (9)

$$\chi(x) = \begin{cases} 1, x < 0 \\ 0, x \geq 0 \end{cases}$$ (10)

where $d_{ij}$ is the Euclidean distance of data point $i$ and $j$, $d_c$ is a parameter which makes the mean value of $\rho_i$ equal to $2\% N$. $N$ is the total number of data points. Larger $\rho_i$ means the point $i$ is more likely to be the center of a cluster.

(2) Computation of $\delta_i$.

$$\delta_i = \max_{j : \rho_j > \rho_i} (d_{ij})$$ (11)

If $\rho_i$ is a global max, let $\delta_i = \max_j (d_{ij})$. $\delta_i$ is the minimum distance between the points which have bigger $\rho$ than $\rho_i$ and the point $i$, a large value of $\delta_i$ means the point $i$ is far away from other dense area.

(3) Drawing the decision figure.

Using $\rho$ as the horizontal axis and $\delta$ as the vertical axis, a decision figure of all the data points can be drawn. Then those points with large product values of $\rho * \delta$ will be the cluster centers.

## 2.6. The Complete Proposed F-matrix Estimation Algorithm

Given any image pair, we can use the SIFT algorithm to first find the $N$ pairs of corresponding feature points $m_n = (x_n, y_n, 1)^T$ and $m_n = (x_{n'}, y_{n'}, 1)^T, (n = 1, 2, ..., N)$. Then the complete F-matrix estimation algorithm can be summarized as follows:

(Step 1) Use the $N$ pairs of points to construct the 4D vectors $q_n = (x_n, y_n, x_{n'}, y_{n'})^T$.
(Step 2) Compute of the local density $\rho_i$ using (9) and (10),
where $d_{ij}$ is now the Euclidean distance between vectors $q_i$ and $q_j$.
(Step 3) Compute the distance $\delta_i$ using (9) and (11).
(Step 4) Obtain the decision figure and find the cluster centers, which are treated as inliers.
(Step 5) Use RANSAC on the inliers of step 4, the F-matrix of the randomly chosen points is computed by the 8-Point method.

As will be shown later, because the clustering step can generally produce more reliable inliers than the traditional methods, the final estimate of the F-matrix will be more accurate. In addition, because of the same reason, less iterations are needed, which means faster running time. We pick the point whose $\delta_i * \rho_i$ is lower than $\alpha * \delta_{max} * \rho_{max}$ as an outlier. Point $i$ whose $\delta_i * \rho_i$ is higher than that is decided as a clustering center in [7]. It represents the basic transformation of the local area. $\alpha$ affects the total number of outliers which are picked out that larger $\alpha$ will result in more outliers. Meanwhile, $\alpha$ also influences the reliability of the inliers. With a large amount of points and $g$ vectors located among an inlier $q_i$, it usually indicates that $g_i$ is quite close to the local mean value and represents the actual transformation between two images at the point $i$.

As is described above, feature points often lie in areas that have obvious characteristics. In a static scene, we consider that these dense points are often eliminated from false matching and are more reliable than the points that are isolated when being used in estimating the F-matrix because of the presence of noise. Besides, the coordinates of the matching points which are corresponding to the same object in 3D space and are located in the same area in the image plane, often change slowly along with the position in image plane. If there is noise whose mean value is 0, we can refer to the other points in the same area to decrease the influence of noise.

In the decision figure, a false corresponding pair $q_i$ has big $\delta_i$ and small $\rho_i$ in common.

The center of a cluster of $q$ has big $\delta$ and $\rho$. The points in a cluster excluding its center have small $\delta_i$ but big $\rho_i$.

As is described in [8], inliers that all come from a small area make no sense because they can hardly represent the transformation between the image pairs in the whole range of the image. However, our optimization method easily overcomes this by choosing the inlier point with a large $\delta$.

## 3. EXPERIMENTS

Ten pairs of images (5 pairs of real images and 5 pairs of synthetic images) captured from ten scenes are used. Then the F-matrix is estimated utilizing the proposed method. We compare the proposed method with the classical methods which are completed by directly calling the functions in OpenCV. We only show the results of the estimation of F-matrix of one pair of real images and one pair of synthetic images here. While the results of the proposed method in the ten pairs of images are all encouraging.

### 3.1. Real Images

When $th = 2.2$, experimental results of the two images captured by a camera are shown in Fig. 4 which is the decision figure in which the curve represents $\alpha = 0.011$ and the bigger circles are outliers found by the conventional RANSAC method. Besides, Fig. 5 shows the matching pairs which are inliers found by the proposed method.

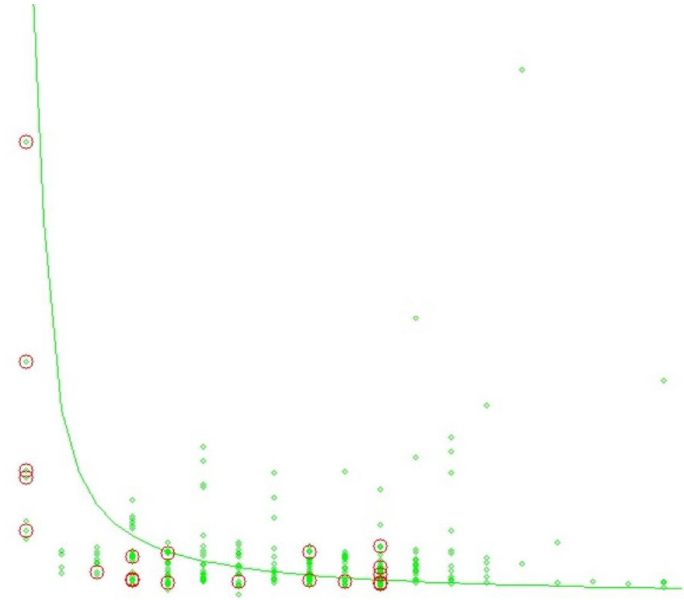

Figure 4.  Decision figure of the pair of real images.

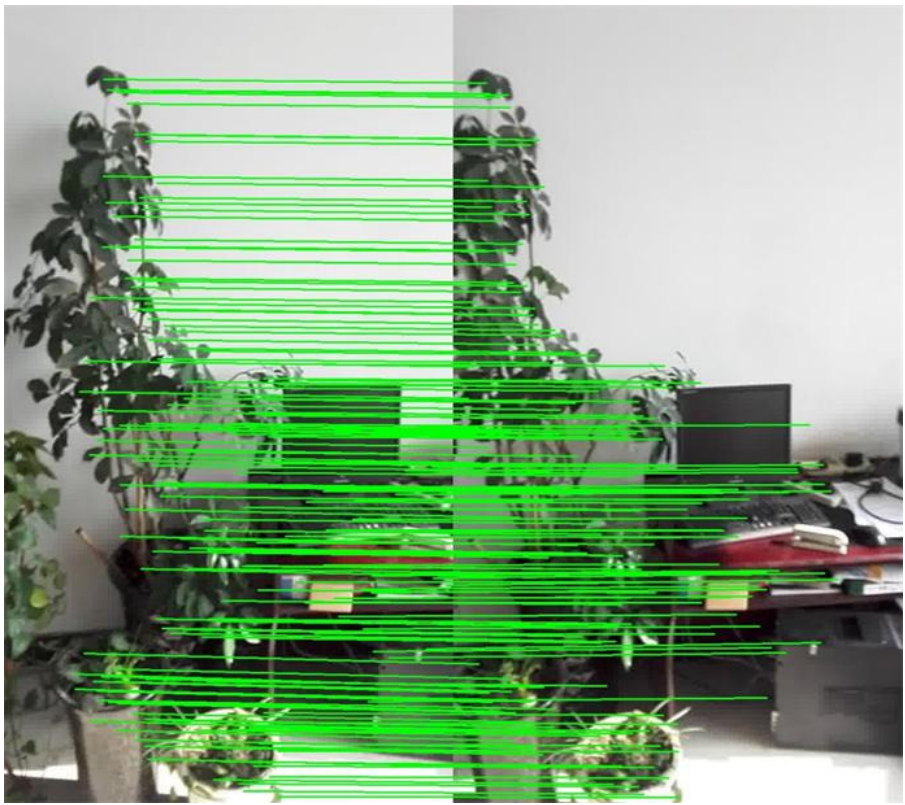

Figure 5.  Matching points of the pair of real images.

As is seen in Fig. 4, most of the outliers in RANSAC (displayed by bigger circles) are under the curve, it proves the reliability of the search of density peaks.

Using different values of $th$, we compare the proposed method with RANSAC whose result is shown in Table 1.

Table 1.  Comparison with RANSAC.

| th(pixel) | Method | Time(ms) | Mean Error(pixel) |
|---|---|---|---|
| 2.2 | RANSAC | 4.7 | 0.841559 |
| | Proposed, $\alpha = 0.011$ | 9.9 | 0.744619 |
| 1 | RANSAC | 10.3 | 0.438181 |
| | Proposed, $\alpha = 0.02$ | 13.3 | 0.412330 |
| 0.8 | RANSAC | 33.3 | 0.370029 |
| | Proposed, $\alpha = 0.025$ | 14.3 | 0.357731 |
| 0.5 | RANSAC | 126 | 0.241158 |
| | Proposed, $\alpha = 0.04$ | 17.2 | 0.224982 |

When $th$ decreases, the time RANSAC consumes increases rapidly because Ransac picks up data points randomly and iterates until a fine F-matrix is found. However, the proposed method has a prior analysis on the characteristics of the data points using a simple algorithm. Beyond that, the mean error in pixel of the proposed method is smaller.

Comparison results with other methods are show in Table 2.

Table 2.  Comparison with other methods.

| Method | Time(ms) | Mean Error(pixel) |
|---|---|---|
| 8-POINT | 1.3 | 1.064726 |
| 7-POINT | 23.6 | 1.706590 |
| LMeDS | 18.2 | 0.833023 |
| Proposed | 9.9 | 0.744619 |

## 3.2. Synthetic Images

To compare the methods from another point of view deeply, we use OpenGL to generate 2 images $I$ and $I'$ which have a known F-matrix $F_0$. So $F_0$ is the ground truth F-matrix. Then we utilize the evaluation method in section 4.1 of [8] proposed by Zhang to evaluate the F-matrix estimated by the methods described above. The evaluation method is computed following the steps (see Fig. 6):

(1) Compute the fundamental matrix $F_1$ the proposed method.

(2) Choose a point $m_i$ from image $I$ and compute the following epipolar lines:

$$l_{0i} = F_0 m_i \tag{12}$$

$$l_{1i} = F_1 m_i \tag{13}$$

(4) Choose a point $m_i'$ on the epipolar line $l_{0i}$ and compute the following epipolar lines:

$$l_{0i}' = F_0^T m_i' \tag{14}$$

$$l_{1i}' = F_1^T m_i' \tag{15}$$

(4) If the epipolar lines do not intersect the image plane, go back to step 2.

(5) Compute the distance $d'_{1i}$ from the point $m'_i$ to the line $l_{1i}$ (c.f. Fig. 6).

(6) Compute the distance $d_{1i}$ from the point $m_i$ to the line $l_{1i}$ (c.f. Fig. 6).

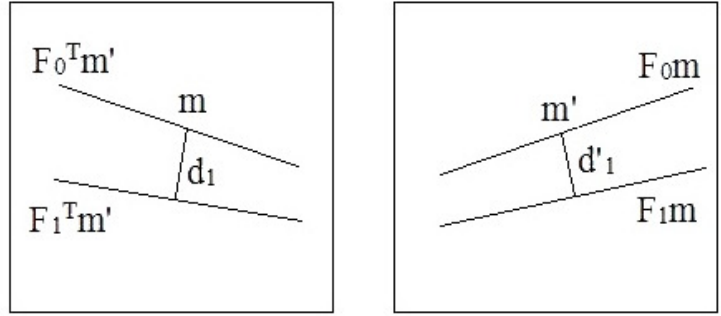

Figure 6. Illustration of the evaluation method [8].

(7) Repeat step 2 to step 6 for $N$ times.

(8) Compute the mean value $d_1$ of all the $d_{1i}$ and $d'_{1i}$.

Then the results that the proposed method compares with other methods are shown in Fig. 7, Fig. 8 and Table 3. The numbers in Table 3 represent the difference between the F-matrix estimated by the methods above and the ground truth $F_0$ while $th = 2.2, \alpha = 0.011$.

Table 3. F-matrix estimation error comparison with ground truth.

|         | 8-POINT   | 7-POINT   | LMeDS    | RANSAC   | Rroposed |
|---------|-----------|-----------|----------|----------|----------|
| Error   | 83.875023 | 42.706934 | 1.300233 | 1.330434 | 1.077631 |

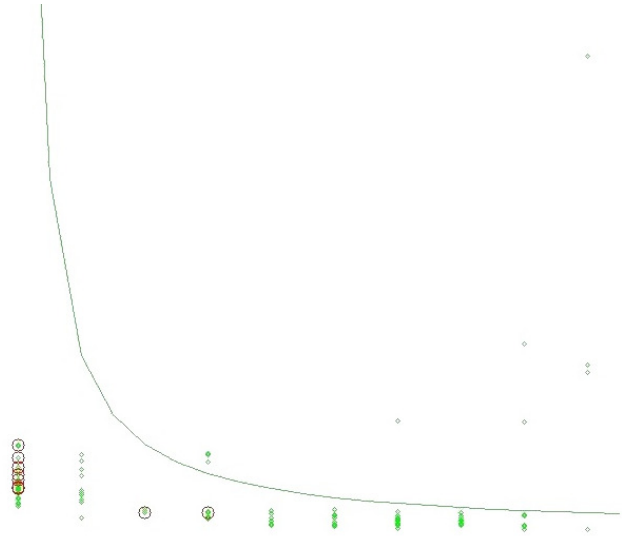

Figure 7. Decision figure of the pair of real images.

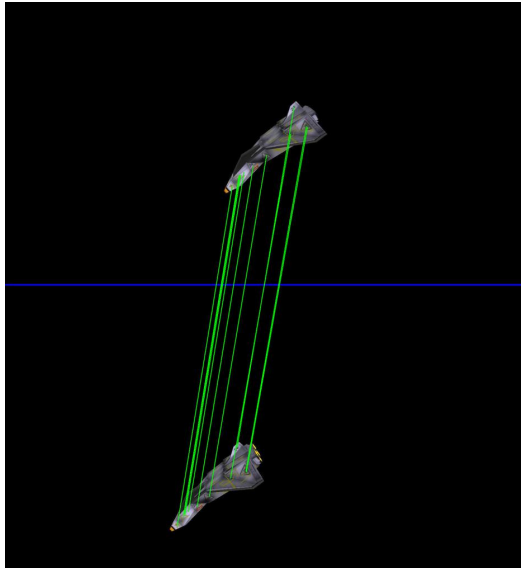

Figure 8. Matching points of the pair of real images.

## 4. CONCLUSION

The paper proposes the utilization of clustering to optimize the corresponding pairs of points for accurate estimation of the fundamental matrix. This approach chooses the point pairs which are likely to represent the true relationship between image pairs and suffer less from noise. As the threshold decreases, the method we use is better than the conventional RANSAC both in processing speed and accuracy. Besides, our optimization based on search of density peaks has a lower level of the error between the estimated F-matrix and the ground truth.

## AUTHORS

Hao Wu was born on Oct. 16, 1988, in Baoji, China. He got bachelor's degree of electronic information technology and instrument in Zhejiang University in 2007. He is studying for a master's degree in Lanzhou University. His interest of research is compute vision.

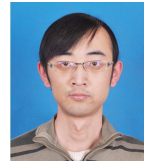

Yi Wan received his Ph.D degree from the Rice University in 2002 and is currently a faculty member at the school of information science and engineering, Lanzhou University, China.

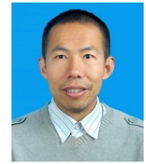